\documentclass[letterpaper, 10 pt, journal, twoside]{IEEEtran}
\usepackage{listings}
\usepackage{xcolor}
\usepackage{amsmath,amssymb,amsfonts}
\usepackage{graphicx}
\usepackage{cite}
\usepackage{hyperref}
\usepackage{color}
\usepackage{url}

\title{Robot Context Protocol (RCP): A Runtime-Agnostic Interface for Agent-Aware Robot Control}

\author{
Lambert Lee$^{1}$, Joshua Lau, 
\thanks{Manuscript received Month DD, YYYY; Accepted Month DD, YYYY.}
\thanks{$^{1}$RoboStack Research Group}%
}

\begin{document}
\maketitle

\begin{abstract}
The Robot Context Protocol (RCP) is a lightweight, middleware-agnostic communication protocol designed to abstract robotic system complexity and enable seamless interaction between robots, users, and autonomous agents. RCP provides a unified and semantically meaningful interface that decouples client-facing operations from backend implementations, supporting a wide range of deployment environments—including physical robots, cloud-based orchestrators, and simulated platforms. Built on HTTP and WebSocket transport layers, the protocol defines a schema-driven message format with structured operations such as read, write, execute, and subscribe. It integrates features such as runtime introspection, asynchronous feedback, multi-tenant namespace isolation, and strict type validation to ensure robustness, scalability, and security. This paper presents a comprehensive overview of RCP’s architecture, message structure, interface model, and adapter-based backend integration strategy. We also outline deployment practices and discuss its applicability across industries including manufacturing, logistics, and healthcare. RCP enables intelligent, resilient, and safe robotic operations in complex, multi-agent ecosystems.
\end{abstract} 
\bigskip
\section{Introduction}
Modern robotic systems have evolved into highly sophisticated platforms, integrating a diverse array of sensors, actuators, and computational modules. These heterogeneous components enable the execution of complex and adaptive behaviors in dynamic environments\cite{1}. However, as the internal architecture of robotic systems becomes increasingly intricate, interfacing with them—especially across differing runtime environments—presents substantial challenges. Conventional integration strategies often necessitate in-depth knowledge of specific middleware frameworks, message-passing systems, and hardware-dependent implementations. This complexity imposes a steep learning curve not only for human developers and researchers, but also for AI agents attempting to access or control robotic functions programmatically.

To mitigate these barriers, we introduce the Robot Context Protocol (RCP), a lightweight, runtime-agnostic communication framework designed to abstract internal system complexity while providing a unified and standardized interface for interaction. RCP enables external systems—whether cloud services, local applications, or autonomous agents—to seamlessly observe robot states, query sensor data, dispatch control commands, and subscribe to event streams. Crucially, these capabilities are accessible without requiring prior knowledge of the robot's underlying middleware, control software, or runtime configuration.

While conceptually aligned with prior work such as rosbridge\cite{7}, which exposed robotic APIs via web protocols, RCP distinguishes itself through a more rigorously layered protocol architecture. It is built atop standardized communication mechanisms such as HTTP for request-response interactions, WebSocket\cite{3} for real-time bi-directional messaging, and Server-Sent Events (SSE) for lightweight, unidirectional data streaming. This tri-channel design not only enhances modularity and extensibility, but also ensures compatibility with both cloud-based robotic platforms and physically deployed robotic systems. By decoupling communication logic from execution details, RCP supports both agent-to-robot (A2R) and human-to-robot (H2R) interaction models, serving as a foundational interface for AI-augmented robotic ecosystems.

This paper is organized as follows. We begin in Section2 by presenting the overall protocol architecture, including the layered design comprising the Adapter Layer and Transport Layer. Section3 introduces the schema-driven message format employed by RCP, describing the structure of requests and responses, supported data types, and interaction semantics. In Section4, we detail the unified context model and namespacing strategy that abstract low-level system resources into addressable interfaces. Section5 focuses on the feedback and status mechanisms that enable structured reporting of command outcomes and runtime health. Section6 outlines the robustness features and deployment strategies that support scalable integration in production environments. Finally, Section7 concludes with a summary of key contributions and a discussion of future directions for protocol evolution and system integration.

\section{Definition of the Robot Context Protocol}

Fundamentally, RCP operates as a context abstraction layer, mediating between robotic systems and external users—ranging from conventional web clients to large language model (LLM)-based autonomous agents—without disclosing internal control loops, execution threads, or device-specific configurations. The protocol encapsulates all interaction capabilities into a cohesive set of high-level application programming interfaces (APIs). These include operations for retrieving live sensor data, executing movement or actuation commands, and subscribing to system-level events such as mode transitions, fault states, or behavior completions.

This abstraction not only simplifies the development of robotics-aware applications, but also enables RCP to function seamlessly across diverse deployment contexts—including on-device, edge-based, or cloud-native infrastructures. In doing so, RCP allows robotic systems to be treated as modular, context-aware services within broader multi-agent networks, thereby promoting interoperability, scalability, and ease of access in next-generation human-AI-robot collaborations.

Communication within the RCP is structured around two primary mechanisms: HTTP endpoints and WebSocket channels\cite{3}. These two modes of interaction are designed to cover both synchronous, stateless operations and asynchronous, real-time data exchange.

The \textbf{HTTP interface} supports structured, stateless requests that are ideal for discrete transactions, such as querying robot states, retrieving parameter values, or invoking one-shot control commands. These endpoints follow RESTful principles and are accessible via standard web client calls, facilitating integration with conventional cloud services and frontend applications.

The \textbf{WebSocket interface}, on the other hand, establishes a persistent, bidirectional channel that enables low-latency, real-time communication. This mode is particularly well-suited for event streaming, continuous telemetry monitoring, and interactive control loops. Through these two channels, RCP exposes a unified interface to all robot-readable and writable entities—including topics, parameters, actions, and services—irrespective of the robot’s underlying middleware or runtime system.

The implementation of RCP adheres to a layered architectural paradigm that cleanly separates system responsibilities and facilitates modularity. At the core of this design is the \textbf{Message Transformation Layer}, which orchestrates communication between external clients and the robot’s native control stack. This layer is further subdivided into two principal components: the \textbf{Adapter Layer} and the \textbf{Transport Layer}.
\begin{figure}[t]
  \centering
  \includegraphics[width=\linewidth]{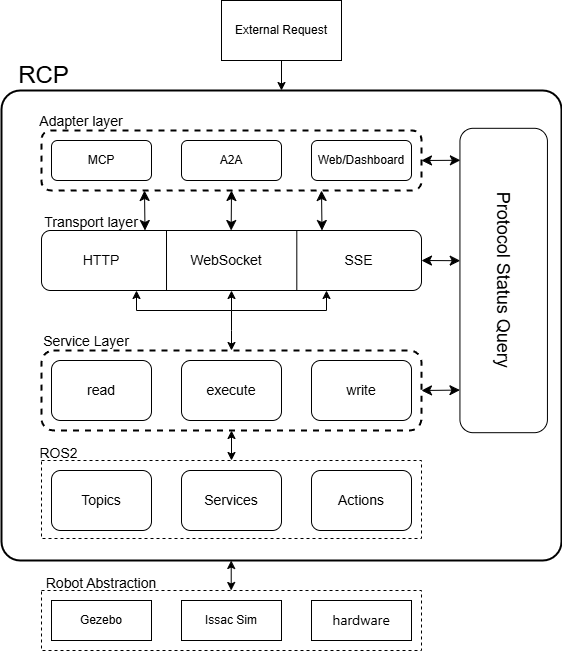}
  \caption{Overall architecture of the Robot Context Protocol (RCP). The system is structured in modular layers: the Adapter Layer translates diverse client interfaces (e.g., MCP, A2A, dashboards) into unified RCP requests; the Transport Layer handles communication via HTTP, WebSocket, and Server-Sent Events (SSE); the Service Layer abstracts commands into high-level operations—read, execute, and write; and the ROS2 Interface Layer maps these operations onto native ROS2 constructs such as Topics, Services, and Actions. A dedicated status query path provides protocol and adapter health diagnostics. The RCP stack can interface seamlessly with physical robots, simulators like Gazebo or Isaac Sim, or cloud-hosted deployments.}
  \label{fig:rcp-architecture}
\end{figure}

\section{Protocol Architecture}
The Robot Context Protocol (RCP) is structured as a multi-layered architecture that facilitates modular and scalable communication between external agents and robotic systems. Each layer is designed to abstract complexity, promote interoperability, and support deployment across a wide range of operational contexts.

At the highest level, the \textbf{Adapter Layer} enables seamless integration with heterogeneous client types, including large language models (via MCP), autonomous agents (via A2A), and human interfaces such as dashboards. These adapters normalize diverse interaction modalities into a unified message format.

Following adaptation, messages are processed by the \textbf{Transport Layer}, which provides support for three communication mechanisms: HTTP for synchronous request-response transactions, WebSocket for real-time bidirectional streaming, and Server-Sent Events (SSE) for efficient unidirectional message broadcasting. This tri-channel design supports low-latency, reliable communication for a variety of application scenarios.

The \textbf{Service Layer} encapsulates protocol functionality through a minimal and expressive set of operations—\texttt{read}, \texttt{write}, \texttt{execute}, and \texttt{subscribe}. This layer abstracts the control surface of the robot into a semantically coherent interface, enabling uniform access to system capabilities.

Beneath the service interface, the \textbf{ROS2 Interface Layer} serves as an abstraction envelope that translates high-level commands into ROS2-native constructs such as topics, services, actions, and parameters. This mapping is performed in a type-safe, schema-driven manner that decouples external access from internal middleware conventions.

Complementing these layers is a \textbf{Status and Monitoring Module} that exposes real-time protocol health, adapter readiness, and diagnostic metadata. This capability supports runtime introspection, autonomous fault handling, and system-level orchestration.

This layered protocol design enables RCP to function as a runtime-agnostic control interface, capable of bridging physical hardware, simulation environments, and cloud-native robotic platforms. An overview of the system architecture is depicted in Fig.~\ref{fig:rcp-architecture}.

\paragraph{Adapter Layer}  
The Adapter Layer is tasked with handling user-specific logic, converting diverse forms of user inputs into protocol-compliant RCP messages. It provides an extensible mechanism for integrating heterogeneous interaction modalities. Currently implemented adapters include:

\begin{itemize}
\item \textbf{MCP Adapter:} Translates outputs generated by LLMs into structured command messages, enabling natural language interaction with robotic systems.

\item \textbf{A2A Adapter:} Facilitates agent-to-agent coordination by converting symbolic planning outputs into executable task sequences suitable for robotic execution.

\item \textbf{Web/Dashboard Adapter:} Supports graphical interfaces and dashboard applications, offering RESTful endpoints and visualization-ready telemetry.
\end{itemize}

This modular approach ensures that new client types—such as mobile applications, command-line tools, or gRPC-based microservices—can be accommodated by implementing additional adapters, without requiring changes to the underlying transport logic or core RCP message format.

\paragraph{Transport Layer}Following adapter processing, messages are relayed through the Transport Layer, which standardizes the formatting, delivery, and routing of RCP messages. This layer serves as the protocol’s interface to external systems:

\begin{itemize}
\item The \textbf{HTTP interface} manages short-lived, synchronous operations, including data queries, service invocation, and configuration updates.

\item The \textbf{WebSocket interface} supports long-lived sessions for use cases that demand real-time responsiveness, such as continuous data streaming, subscription-based event updates, and live task monitoring \cite{3}.

\item The \textbf{Server-Sent Events (SSE) interface} provides a unidirectional, lightweight push mechanism ideal for broadcasting periodic status updates or event notifications to browser-based or resource-constrained clients. SSE is well-suited for applications where WebSocket overhead is unnecessary or unsupported.
\end{itemize}

Together, these communication channels enable RCP to deliver low-latency, scalable, and interoperable interactions for both human operators and autonomous agents.

The layered communication model implemented in RCP ensures consistency across all external interactions. Incoming and outgoing messages undergo uniform validation and schema enforcement, and are tagged with session metadata for robust tracking and error handling. By abstracting internal system details, RCP allows external users to interface with robots in a predictable, secure, and implementation-agnostic manner.

This unified architecture enables seamless operation across local deployments, cloud environments, or hybrid configurations. As a result, RCP serves as a foundational layer for building agent-compatible, context-aware robotic ecosystems that can be accessed and orchestrated by a wide range of intelligent systems and human operators alike.

\paragraph{Status Query (RCP Internal Health and Command Feedback)} 
To support reliability, observability, and operational transparency, the Robot Context Protocol (RCP) incorporates a dedicated \textbf{status feedback system}. This system plays a central role in ensuring that external users and agents can make informed decisions about robot state, system availability, and error conditions without relying on internal diagnostics or middleware-specific monitoring tools.

The status system continuously tracks and exposes a range of runtime metrics that characterize the health and availability of both the protocol infrastructure and its connected adapters. These metrics include, but are not limited to: system uptime, memory and CPU usage (where available), adapter readiness, backend connectivity, message queue backpressure, and recently logged warnings or errors. This data is periodically refreshed and made accessible via a standardized status query endpoint, which can be polled or subscribed to depending on the client’s communication mode \cite{5}.

Clients—whether human operators, graphical interfaces, or autonomous agents—can issue status queries to verify that RCP is operational, determine the connection status of core adapters (such as MCP for natural language interaction or A2A for multi-agent coordination), and inspect whether any command execution faults have occurred. This capability is particularly important in fault-tolerant systems, distributed control environments, or scenarios involving dynamic orchestration of multiple robotic agents, where awareness of system state directly informs execution logic and contingency planning.

In addition to reporting passive system health, RCP also delivers \textbf{structured command-level feedback} for every operation initiated through the protocol, regardless of the communication channel used (HTTP, WebSocket, or adapter-mediated). Upon command submission, the system returns a detailed status response encapsulating the outcome of the operation. This response specifies whether the command has been accepted, successfully executed, rejected, or marked as pending due to delayed fulfillment or dependency resolution \cite{2,3}.

Each status message is enriched with semantic annotations and, when applicable, human-readable diagnostic text that clarifies the reason for the result. This includes messages such as command accepted,'' execution completed,'' resource unavailable,'' parameter out of bounds,'' or ``action in progress.'' In error scenarios, the system may also attach diagnostic metadata such as the error type, origin module, recommended remediation steps, or stack trace references (if supported by the backend).

Illustrative examples of status responses include:

\begin{itemize}
  \item \texttt{Command /action/move\_to executed successfully.}
  \item \texttt{Failed to apply configuration: parameter 'speed\_limit' exceeds allowed range.}
  \item \texttt{Warning: action '/navigate\_to' is currently in progress; rejecting duplicate request.}
  \item \texttt{Command rejected --- MCP adapter is not connected to the runtime.}
\end{itemize}

This structured feedback mechanism is critical for implementing reliable higher-order logic in external systems. For instance, agents can automatically retry failed operations, initiate fallback actions, or escalate to human oversight when critical faults are detected. It also simplifies the design of user interfaces by providing immediate, interpretable feedback that can be displayed to operators without revealing internal protocol or middleware details.

Furthermore, RCP's feedback mechanism is protocol-consistent across all endpoints and interaction modalities. This uniformity ensures that whether a command is submitted by a RESTful web client, a WebSocket subscriber, or a language model agent, the resulting status information follows a predictable schema and semantic pattern.
\begin{figure}[t]
  \centering
  \includegraphics[width=\linewidth]{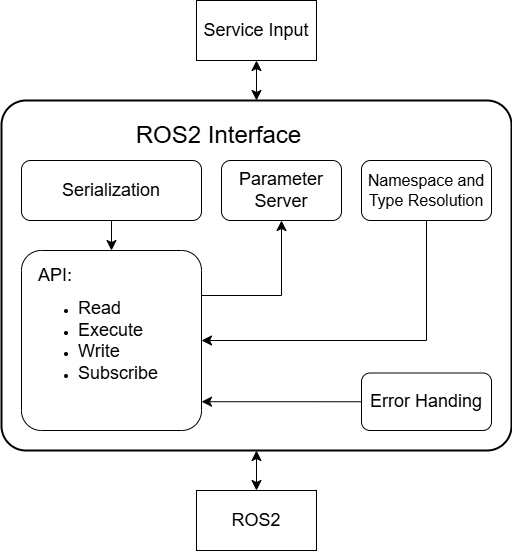}
  \caption{Architecture of the ROS2 Interface Layer within the Robot Context Protocol (RCP). This component mediates between high-level RCP service requests and the underlying ROS2 runtime. The API handler supports standardized operations—read, execute, write, and subscribe—while the serialization module ensures payload compatibility with ROS2 message types. The parameter server provides access to runtime configuration values, enabling dynamic reconfiguration. This layered abstraction allows RCP to expose a unified control surface without revealing internal ROS2 constructs to external agents.}
  \label{fig:ros-architecture}
\end{figure}
\paragraph{ROS2 Operations Interface}
At its core, RCP encapsulates the complexity of the underlying ROS2 runtime through a semantically structured interface layer that presents a unified, middleware-agnostic control surface. This interface abstracts native ROS2 constructs—including topics, services, actions, and parameters—into standardized RCP operations such as \texttt{read}, \texttt{write}, \texttt{execute}, and \texttt{subscribe}. Internally, these operations are mapped through handler modules that bridge to ROS2 primitives, enabling external clients to issue commands or retrieve data without direct exposure to ROS2-specific message types, service signatures, or node architecture \cite{5,8}.

The design draws from the DDS-based publish-subscribe paradigm of ROS2, while deliberately abstracting low-level details such as Quality-of-Service (QoS) settings, discovery protocols, and type negotiation \cite{6}. This abstraction allows for a simplified, developer-friendly interface that supports integration across varied deployment environments, including embedded systems, high-fidelity simulators, and cloud-hosted platforms.

A distinctive enhancement introduced by RCP is its native support for multi-tenant communication. Clients operate within logically isolated namespaces—e.g., \texttt{/tenant/alpha} or \texttt{/tenant/beta}—which scope access to sensor data, actuators, configuration parameters, and system actions. This namespacing mechanism enables concurrent control by multiple agents while maintaining secure and context-aware separation of state, operations, and privileges.

An overview of this abstraction model and its integration with ROS2 runtimes is illustrated in Fig.~\ref{fig:ros-architecture}.

Recommended command syntax for interacting in a multi-tenant environment might include:
\begin{itemize}
  \item \texttt{read /tenant/alpha/sensor/pose} — Query position within tenant \texttt{alpha}
  \item \texttt{execute /tenant/beta/action/move\_to} — Command navigation for tenant \texttt{beta}
  \item \texttt{write /tenant/alpha/param/speed\_limit} — Adjust motion constraint in context of \texttt{alpha}
\end{itemize}

By combining runtime transparency, structured feedback, and system health monitoring, RCP enables external systems to engage with robotic agents in a resilient, introspective, and backend-agnostic manner. This design is critical for enabling long-running autonomous operation, collaborative multi-agent systems, and intelligent user interfaces that adapt to the real-time status of the robotic platform.

The abstraction model implemented by the RCP encompasses a broad range of robotic operation domains, offering a unified interface for both human and agent-level interaction. These domains include but are not limited to: continuous streaming of sensor data (e.g., camera images, LiDAR point clouds, IMU readings), dispatching motion commands for navigation and actuator control, invoking internal diagnostic or maintenance services, and adjusting runtime parameters such as velocity caps, sensor gains, or behavior modes.

Each of these functions is exposed using a consistent path-based addressing scheme, designed to decouple clients from backend implementation details. Paths are organized semantically to reflect the logical structure of the robot's capabilities. Illustrative examples include:
\begin{itemize}
  \item \texttt{/sensor/pose} — Access robot position and orientation
  \item \texttt{/action/move\_to} — Trigger navigation to a target location
  \item \texttt{/param/speed\_limit} — Modify motion constraints
  \item \texttt{/service/reset\_system} — Invoke system reboot or fault recovery
\end{itemize}

These abstract paths are designed to be accessed using a small, standardized set of operations: \texttt{read}, \texttt{write}, \texttt{execute}, and \texttt{subscribe}. This interaction model enables clients to perform queries, updates, and control operations in a consistent manner regardless of backend complexity.

For example, rather than requiring clients to subscribe to backend-specific topics such as \texttt{/odom} or \texttt{/robot\_pose\_ekf/odom\_combined}, a request to \texttt{read /sensor/pose} retrieves the robot's current estimated position\cite{5}. Likewise, issuing \texttt{execute /action/move\_to} eliminates the need for ROS2-specific action interfaces like \texttt{/move\_base}\cite{5}.. To modify configuration parameters—such as updating the robot’s maximum speed—clients can use \texttt{write /param/speed\_limit}, without interacting directly with ROS2’s parameter services or dynamic reconfigure tools\cite{5}. A full system reset can be initiated via \texttt{execute /service/reset\_system}, regardless of whether the underlying implementation relies on a service, action, or custom node logic.

This unified interface design confers several key advantages. First, it is environment-agnostic: no ROS2-specific topic names, message types, or service signatures are exposed to external users. Second, it introduces a clean, human-readable, and agent-friendly command syntax, reducing the barrier to integration for developers, researchers, and AI systems. Third, the interface supports runtime introspection, allowing clients to discover all available paths and associated operations dynamically through a discovery API. This promotes adaptability, as clients can reconfigure or evolve without requiring prior hardcoding of system knowledge.

Internally, RCP organizes all robot-accessible resources within a \textbf{unified context model}. This model standardizes representation across sensors, actuators, high-level actions, runtime parameters, and events such as boot cycles, task completions, or fault states. Each resource is defined using a structured namespace and schema-based type signature, facilitating automatic parsing, validation, and error reporting \cite{6}. Clients are able to enumerate this namespace through protocol-level queries, enabling fully dynamic interaction models and adaptive control strategies.

To support the demands of production-grade deployment, RCP integrates several robustness features adapted from prior systems like rosbridge\cite{2,7}. These include:

\begin{itemize}
  \item Emulation of asynchronous services with structured response handling.
  \item Optional message compression for bandwidth-sensitive scenarios.
  \item Segmentation and reassembly for large payloads (e.g., image or map data).
  \item Strict schema enforcement for all messages, ensuring type correctness and protocol conformance.
  \item Persistent session tracking to support long-running agent interactions.
  \item Endpoint-level access control and authentication for multi-tenant environments.
\end{itemize}

Although the RCP protocol surface is fully backend-agnostic, it relies internally on a system of \textbf{robot-facing adapters} to bridge with operational runtimes such as ROS2\cite{5}. Each native capability—whether defined as a topic, service, or action—is wrapped by an RCP handler module. These modules map the backend functionality onto the HTTP/WebSocket protocol endpoints in a consistent and transparent manner, ensuring that clients interact with the robot through a uniform abstraction, regardless of deployment modality.

As a result, RCP enables seamless integration with physical robotic systems, high-fidelity simulators\cite{4}, or cloud-hosted control environments\cite{10,11}, all without requiring modifications to the external protocol. This layered abstraction forms a clean architectural boundary between internal robotic logic and external consumer access, empowering a wide range of use cases from single-user control panels to fully autonomous multi-agent orchestration platforms.

\section{Message Format and Interface Definition}

The Robot Context Protocol (RCP) defines a structured, schema-driven message format to ensure consistent communication between external clients and robotic systems~\cite{2,7}. All interactions—whether over HTTP or WebSocket—adhere to a unified message structure that abstracts low-level middleware details and provides semantic clarity for client applications. The format is designed to be both human-readable and machine-parseable, supporting robust validation, introspection, and interoperability~\cite{3,6}.

\subsection{Message Envelope}

Each RCP message consists of an envelope containing metadata and a body containing the request or response payload. The envelope includes fields such as:

\begin{itemize}
  \item \texttt{type} -- Indicates the operation category (e.g., \texttt{"read"}, \texttt{"write"}, \texttt{"execute"}, \texttt{"subscribe"}, \texttt{"status"}).
  \item \texttt{path} -- Specifies the logical target resource using a namespace-based identifier (e.g., \texttt{/sensor/pose}, \texttt{/param/speed\_limit}).
  \item \texttt{id} -- A client-specified identifier used to track the request-response lifecycle, particularly for asynchronous operations.
  \item \texttt{timestamp} -- (Optional) UTC timestamp for temporal tracking.
\end{itemize}

The message \textbf{body} is context-dependent and varies according to the operation type. For example, a \texttt{write} operation includes a data payload, while a \texttt{read} operation may contain filter parameters or sampling hints.

\subsection{Data Schema and Types}

All payloads in RCP conform to JSON-based schemas defined for each addressable path~\cite{6}. Each path corresponds to a resource with a specified type signature, including:

\begin{itemize}
\item \textbf{Primitive types:} \texttt{int}, \texttt{float}, \texttt{bool}, \texttt{string}
\item \textbf{Compound types:} arrays, dictionaries (maps), and nested structs
\item \textbf{Time types:} ISO-8601 formatted timestamps or UNIX epoch
\item \textbf{Geometry types:} standardized representations for pose, twist, acceleration, etc.
\end{itemize}

For instance, a query to \texttt{/sensor/pose} may return a structured response like:

\begin{lstlisting}
{
  "position": { "x": 1.23, "y": 4.56, "z": 0.00 },
  "orientation": { "x": 0.0, "y": 0.0, "z": 0.0, "w": 1.0 },
  "frame_id": "map",
  "timestamp": "2025-05-29T14:12:04Z"
}
\end{lstlisting}

Each response is strictly schema-validated against its path definition to ensure type safety and compatibility across clients~\cite{6}.

\subsection{Data Abstraction and Namespacing}

All robot-readable and robot-writable data exposed through the RCP are organized under a unified context model, which abstracts low-level system constructs into semantically meaningful objects~\cite{2,5,10}. This abstraction spans a wide range of domains, including sensor data such as outputs from cameras, inertial measurement units (IMUs), and LiDAR devices; actuator commands targeting components like wheels, robotic arms, and grippers; system-level events including boot sequences, idle states, or error notifications; task-level instructions such as navigation or pick-and-place directives; and configuration parameters that define runtime behavior, including velocity limits or controller modes.

Each of these elements is represented by a unique namespace and associated type descriptor, forming a structured and addressable hierarchy~\cite{5,6}. This naming scheme enables dynamic discovery and interaction without requiring external clients to hardcode internal system endpoints. Instead, clients can introspect the available context at runtime using a discovery API, which enumerates the operational interfaces exposed by the robot~\cite{2}. This model provides a consistent and scalable method of exposing robot capabilities to user interfaces, autonomous agents, and cloud orchestration systems~\cite{10,11}.

\subsection{Interface Contracts and Introspection}

RCP supports a \texttt{discovery} endpoint that returns a complete catalog of available paths, each annotated with:

\begin{itemize}
\item Supported operations (e.g., \texttt{read}-only, \texttt{execute}-enabled)
\item Expected input/output schema definitions
\item Human-readable descriptions and usage examples
\end{itemize}

This enables dynamic introspection and client-side code generation~\cite{7,11}, allowing users and agents to adaptively compose valid RCP requests without hardcoding schema knowledge.

\subsection{Asynchronous Semantics and Feedback}

For operations like \texttt{execute} or \texttt{subscribe}, RCP messages support asynchronous feedback via a subscription channel or response event~\cite{6,9}. Clients are notified through response messages with status updates such as:

\begin{itemize}
  \item \texttt{"accepted"} -- The command is queued for execution.
  \item \texttt{"in\_progress"} -- The action has started and is being monitored.
  \item \texttt{"completed"} -- The action completed successfully.
  \item \texttt{"failed"} -- An error occurred; includes diagnostic fields.
\end{itemize}

These updates are identified by their \texttt{id} field to allow the client to correlate responses with original requests.

\subsection{Security and Namespacing}

Each message is scoped to a namespace (e.g., \texttt{/tenant/alpha/}) to support secure multi-tenant operation. The protocol enforces access control lists (ACLs) and authentication at the transport layer to restrict interaction based on user roles or client certificates~\cite{6,10,11}.

\subsection{Extensibility}

Because the RCP message format is schema-defined and transport-neutral, it supports extension via custom fields, version negotiation, and dynamic field registration~\cite{2,6}. Future message versions may include richer metadata, custom events, or alternative encodings (e.g., CBOR or Protobuf) for efficiency in constrained environments~\cite{6}.

In summary, the RCP message format combines semantic clarity with strict structure and validation, allowing heterogeneous systems—including LLMs, user interfaces, and autonomous agents—to communicate with robots in a predictable, extensible, and middleware-agnostic manner.

\section{Robustness and Deployment Features}

\subsection{Robustness and Protocol-Level Features}
To support reliable and scalable integration across real-time and distributed systems, RCP incorporates a set of robust communication features inspired by established frameworks such as \textit{rosbridge}~\cite{2,7}. These include asynchronous service emulation, in which traditional blocking service calls are replaced by non-blocking request-response patterns using event-driven callbacks. This allows external agents to issue commands without stalling, while still receiving feedback on success or failure. Message compression is available to reduce network load, particularly in bandwidth-constrained environments~\cite{7}. Large payloads, such as high-resolution images or volumetric map data, are automatically segmented and reassembled by the protocol stack to ensure transport reliability~\cite{6,7}. All messages exchanged via RCP undergo strict schema validation, which enforces type safety and ensures that malformed or ambiguous data are rejected before reaching backend systems~\cite{6}. Persistent session tracking is implemented to maintain context over long-lived interactions, supporting applications such as continuous monitoring, adaptive learning, or multi-stage task execution~\cite{11}. Additionally, RCP includes security and isolation mechanisms at the endpoint level, allowing developers to define access permissions and enforce multi-tenant policies within shared robot platforms~\cite{10}.

These features collectively enhance the reliability, security, and efficiency of RCP as a middleware-agnostic protocol capable of bridging local and cloud-based robotic infrastructure.

\subsection{Backend Integration and Deployment Flexibility}
Although RCP is designed to be agnostic to the underlying robotics framework, its internal implementation follows an adapter-based integration model~\cite{2,5,8}. Each native robot capability—whether defined as a topic, service, action, or parameter within a framework such as ROS2—is wrapped by a standardized RCP handler. These handlers translate between the internal backend interface and the external protocol layer, exposing their functionality over HTTP and WebSocket endpoints in a consistent and backend-neutral form. This architectural separation allows RCP to operate uniformly across a variety of deployment contexts, including physical robots operating at the network edge, simulated robots hosted in virtualized environments, and large-scale cloud-native robotics orchestration platforms~\cite{10,11}. By decoupling client interaction logic from execution infrastructure, RCP establishes a clean and maintainable boundary between robot runtime systems and external control interfaces, thereby simplifying integration, improving scalability, and facilitating modular system design.

\section{Summary and Future Directions}
The Robot Context Protocol (RCP) presents a unified, transport-agnostic framework for interacting with robotic systems across diverse deployment environments. By abstracting low-level constructs such as ROS2 topics, services, and parameters into a semantically organized context model, RCP enables external agents—human or autonomous—to engage with robots using intuitive, schema-validated interfaces. This simplification reduces integration overhead and fosters interoperability across hardware, simulation, and cloud platforms.

Key features such as message introspection, asynchronous service emulation, dynamic discovery, and namespace-based multi-tenant support establish RCP as a scalable and secure interface layer. Its support for both HTTP and WebSocket transports ensures compatibility with modern web and cloud-native infrastructures, while the adapter-based backend model provides the flexibility to support heterogeneous runtime stacks without altering external APIs.

Together, these design principles position RCP as a robust foundation for next-generation robotics systems—enabling seamless orchestration, cross-platform interaction, and AI-enhanced autonomy. As robotics continues to evolve toward distributed, intelligent, and modular architectures, RCP offers a protocol-level backbone to unify communication, enforce structure, and empower scalable collaboration.

While the Robot Context Protocol (RCP) establishes a robust foundation for decoupled, scalable robot interaction, several avenues remain open for refinement and future enhancement.

First, extending RCP’s support to alternative transport encodings such as CBOR or Protobuf could significantly improve performance in resource-constrained and low-latency environments, particularly in edge or embedded deployments. Similarly, incorporating richer Quality-of-Service (QoS) semantics—akin to DDS in ROS2—would enable finer control over data delivery guarantees, prioritization, and bandwidth usage.

Second, formalizing RCP’s schema registry and message introspection standards will support greater interoperability across vendors and robotic platforms. This could pave the way for a shared ecosystem of open, versioned interface definitions, facilitating toolchains for automated client generation, validation, and simulation.

Third, deeper integration with agent planning systems and foundation models (LLMs, VLMs) can enrich the semantic expressiveness of RCP, allowing high-level goals to be translated into protocol-level actions through natural language or symbolic reasoning. This would further lower the barrier for intelligent agents to control and coordinate robot behavior without rigid task programming.

Finally, future development may explore decentralized governance and event-driven automation through mechanisms such as blockchain-based access control, multi-agent consensus protocols, or context-aware policy enforcement at the protocol layer.

In summary, RCP offers a compelling base for unified robot communication, and its modular design opens the door to a wide range of extensions that will support the next generation of intelligent, distributed, and adaptive robotic systems.
$\,$

$\,$

\bibliographystyle{IEEEtran}

\end{document}